\documentclass[sigconf]{acmart}
\usepackage{algorithm}
\usepackage{caption}
\usepackage{algorithmic}
\usepackage{makecell}
\usepackage{wrapfig}

\usepackage{natbib}
\usepackage{enumitem}
\usepackage{listings}
\usepackage{CJKutf8}
\usepackage{siunitx}
\usepackage{footnote}

\usepackage{fancyvrb}
\usepackage{latexsym}
\usepackage{xspace}
\usepackage[table]{xcolor}
\usepackage{booktabs}
\usepackage{graphicx}
\usepackage{bbding}
\usepackage{amsthm}
\usepackage{hyperref}
\usepackage{url}
\usepackage{thmtools}
\usepackage{mathtools}
\usepackage{upgreek}
\usepackage{dsfont}
\usepackage{cleveref}
\usepackage{bbm}

\usepackage{multirow}
\usepackage{balance}
\usepackage{longtable}
\usepackage{array}
\usepackage[utf8]{inputenc}
 
\usepackage{amssymb}
\usepackage{amsmath}
\usepackage{colortbl}
\usepackage{bm}
\usepackage{microtype}
\usepackage{tcolorbox}
\usepackage{adjustbox}

\AtBeginDocument{%
  }

\copyrightyear{2026}
\acmYear{2026}
\setcopyright{cc}
\setcctype{by-nc-nd}
\acmConference[SIGIR '26]{Proceedings of the 49th International ACM SIGIR Conference on Research and Development in Information Retrieval}{July 20--24, 2026}{Melbourne, VIC, Australia.}
\acmBooktitle{Proceedings of the 49th International ACM SIGIR Conference on Research and Development in Information Retrieval (SIGIR '26), July 20--24, 2026, Melbourne, VIC, Australia}
\acmISBN{979-8-4007-2599-9/2026/07}
\acmDOI{10.1145/3805712.3808505}

\begin{document}

\title{Meituan Merchant Business Diagnosis via Policy-Guided Dual-Process User Simulation}

\author{Ziyang Chen}
\orcid{0000-0002-1714-0304}
\affiliation{
 \institution{LongCat Interaction Team, \\Meituan Inc.}
 \city{Beijing}
 \country{China}
}
\email{chenziyang319@163.com}

\author{Renbing Chen}
\orcid{0009-0000-1660-4675}
\affiliation{
 \institution{LongCat Interaction Team, \\Meituan Inc.}
 \city{Beijing}
 \country{China}
}
\email{chenrb@meituan.com}

\author{Daowei Li}
\orcid{0009-0001-8137-3313}
\affiliation{
 \institution{LongCat Interaction Team, \\Meituan Inc.}
 \city{Beijing}
 \country{China}
}
\email{lidaowei@meituan.com}

\author{Jinzhi Liao}
\orcid{0000-0002-2898-6559}
\affiliation{
 \institution{Independent Researcher}
 \city{Changsha}
 \country{China}
}
\email{jinzhiliao19@163.com}

\author{Jiashen Sun}
\orcid{0009-0001-4722-5456}
\email{sunjiashen@meituan.com}
\author{Ke Zeng}
\orcid{0009-0005-9249-0995}
\email{zengke02@meituan.com}
\affiliation{
 \institution{LongCat Interaction Team, \\Meituan Inc.}
 \city{Beijing}
 \country{China}
}

\author{Xiang Zhao}
\orcid{0000-0001-6339-0219}
\authornote{Corresponding author.}
\affiliation{
 \institution{Independent Researcher}
 \city{Changsha}
 \country{China}
}
\email{xiangz@aliyun.com}

\renewcommand{\shortauthors}{Ziyang Chen et al.}

\begin{abstract}
Simulating group-level user behavior enables scalable counterfactual evaluation of merchant strategies without costly online experiments. However, building a trustworthy simulator faces two structural challenges. First, information incompleteness causes reasoning-based simulators to over-rationalize when unobserved factors such as offline context and implicit habits are missing. Second, mechanism duality requires capturing both interpretable preferences and implicit statistical regularities, which no single paradigm achieves alone. We propose Policy-Guided Hybrid Simulation (PGHS), a dual-process framework that mines transferable decision policies from behavioral trajectories and uses them as a shared alignment layer. This layer anchors an LLM-based reasoning branch that prevents over-rationalization and an ML-based fitting branch that absorbs implicit regularities. Group-level predictions from both branches are fused for complementary correction. We deploy PGHS on Meituan with 101 merchants and over 26,000 trajectories. PGHS achieves a group simulation error of 8.80\%, improving over the best reasoning-based and fitting-based baselines by 45.8\% and 40.9\% respectively.
\end{abstract}



\maketitle

\section{Introduction}

\begin{figure*}[htbp!]
    \centering
    \includegraphics[width=0.9\linewidth]{model.png}
    \caption{Overview of the PGHS framework. \textbf{Phase A} abstracts decision policies through macro-level persona clustering and micro-level rationale refinement. \textbf{Phase B} produces predictions via semantic reasoning and statistical fitting, both conditioned on the abstracted policies. \textbf{Phase C} aggregates dual-process outputs into a group-level estimate via population-aligned sampling.}
    \label{fig:framework}
\end{figure*}

Merchant business diagnosis on large-scale platforms requires answering counterfactual questions about operational strategies~\citep{joyce1999foundations,reymen2015understanding,pearl2009causal,DBLP:journals/corr/abs-2510-13291}. For example, platform operators need to estimate whether adjusting pricing, restructuring menus, or redesigning promotional pages would improve a merchant's conversion rate before committing to costly changes. Since each merchant serves a distinct user population with heterogeneous preferences, such diagnosis demands accurate simulation of group-level user behavior under hypothetical interventions. Online A/B testing remains the gold standard for causal evaluation~\citep{DBLP:conf/kdd/YinH19,DBLP:conf/www/YuanAK21}, but it suffers from critical scalability limitations. Deploying unproven strategies risks user experience degradation and financial loss~\citep{DBLP:conf/vldb/SilversteinBMU98,DBLP:conf/cikm/KasugaY24}. Modern strategies also involve high-dimensional semantic variables such as promotional copywriting and visual layouts~\citep{eberhardt2007causation,hoffman2025step}, creating a combinatorial space that physical experiments cannot cover~\citep{basse2018limitations}. These constraints motivate simulation-based counterfactual evaluation~\citep{DBLP:conf/cikm/KasugaY24}. While individual choices are stochastic, the collective preferences of homogeneous user segments exhibit stable regularities, making group-level simulation a natural granularity for merchant diagnosis.

Building a trustworthy group-level simulator faces two structural challenges. The first is \textit{information incompleteness}. Real decisions are influenced by unobserved factors such as offline context and implicit habits, yet reasoning-based simulators rely on explicit inputs and tend to produce over-rationalized predictions~\citep{wang2021causal,adornetto2025generative}. User simulation has evolved from rule-based and statistical approaches~\citep{balog2025user,fischer2001user} to LLM-driven agents~\citep{gao2024large,zhang2025llm}, but these agents often behave as hyper-rational decision-makers that fail to replicate bounded rationality. The second challenge is \textit{mechanism duality}. Group behavior simultaneously reflects interpretable preference structures and implicit statistical regularities~\citep{kahneman2011thinking}, and a single paradigm typically excels at only one end of this spectrum. Fitting-based simulators built on representation learning~\citep{cheng2016wide,kang2018self,zhou2018deep} capture implicit population-level regularities but degrade under long-tail sparsity and semantic shifts. Despite their complementary strengths, effectively bridging the reasoning and fitting paradigms remains an unexplored challenge. Naive ensembling falls short because the two operate with fundamentally different inductive biases, and unconstrained fusion can easily introduce drift without a shared mechanism to align them.

We propose Policy-Guided Hybrid Simulation (PGHS), a dual-process framework for robust group-level counterfactual evaluation. PGHS mines transferable decision policies from behavioral trajectories to form an explicit mechanism alignment layer, grounding both branches with domain-specific empirical logic~\citep{Guo2024ControlAgentAC,Li2024FormalLLMIF}. This layer anchors two complementary branches inspired by dual-process cognition~\citep{kahneman2011thinking}. A semantic reasoning branch uses policies to constrain LLM inference and prevent over-rationalization. A statistical fitting branch uses policies as conditional priors to absorb implicit regularities. Predictions from both branches are fused under the shared policy space for complementary correction.

Our contributions are threefold.
\begin{itemize}
    \item We identify information incompleteness and mechanism duality as two structural barriers to trustworthy group-level user simulation for merchant diagnosis.
    \item We propose mining decision policies from behavioral trajectories as a shared alignment layer to jointly guide LLM reasoning and statistical fitting.
    \item We deploy PGHS on Meituan with 101 merchants and 26k+ trajectories, achieving 8.80\% group simulation error with 45.8\% and 40.9\% improvements over the best reasoning-based and fitting-based baselines.
\end{itemize}

\section{Method}
\label{sec:method}

Given a merchant scene $S$, an intervention $a$ (e.g., a modified price tier or promotional copy), and a user $u$, let $y(u,S,a)\in\{0,1\}$ denote the binary purchase outcome. Our goal is to predict the group-level purchase rate $\mathbb{E}_{u}[y(u,S,a)]$. As shown in Figure~\ref{fig:framework}, PGHS operates in three phases: (A) policy mining, (B) dual-branch simulation, and (C) group-level fusion.

\subsection{Decision Policy Abstraction}
\label{sec:policy_mining}

This phase distills noisy online behavioral logs into a concise set of transferable decision policies that serve as explicit proxies for the latent mechanism $u^{\text{mech}}$.

\subsubsection*{Trajectory Selection}
We first construct a high-intent dataset $\mathcal{D}_{\text{choice}} = \{(u_i^{\text{prof}}, S_i, y_i)\}_{i=1}^N$ by filtering raw interaction data to retain only sessions where a user actively compares multiple merchants before making a definitive purchase or exit.

\subsubsection*{Hierarchical Policy Discovery}
We employ a coarse-to-fine pipeline. At the macro level, we serialize each user profile $u^{\text{prof}}$ into natural language and map it to a dense embedding $\mathbf{h}_{\text{prof}} = \Psi_{\text{enc}}(u^{\text{prof}})$. K-Means clustering~\citep{macqueen1967some} on these embeddings decomposes the population into $K$ persona groups $\mathcal{P} = \{P_1, \dots, P_K\}$.

Within each persona $P_k$, we leverage an LLM to generate a textual rationale $r$ for each high-intent choice instance $(u, S, y)$:
\begin{equation}
    r = \mathcal{M}_{\text{LLM}}(\text{Explain } y \mid S, u^{\text{prof}}).
\end{equation}
We then embed all rationales into a semantic vector space $\mathbf{v}_i^{\text{rationale}} = \Psi_{\text{enc}}(r_i)$ and apply HDBSCAN~\citep{campello2013density} for micro-level refinement:
\begin{equation}
    \{R_{k,1}, \dots, R_{k,J_k}\}, \mathcal{N}_{k} = \text{HDBSCAN}(\{\mathbf{v}_i^{\text{rationale}} \mid i \in P_k\}),
\end{equation}
where $\{R_{k,j}\}$ are coherent decision sub-groups and $\mathcal{N}_{k}$ are stochastic outliers that we discard.

\subsubsection*{Dual-Modal Policy Representation}
Each refined cluster $R_{k,j}$ is formalized into a Decision Policy with two complementary representations. An \textit{explicit instruction} $T_{k,j}$ is generated by prompting an LLM to summarize the shared decision logic:
\begin{equation}
    T_{k,j} = \mathcal{M}_{\text{LLM}}(\text{Summarize } \{r \in R_{k,j}\}).
\end{equation}
This text contains structured fields including user characteristics, key decision factors, and a decision guide. A \textit{dense semantic vector} $v_{k,j} = \Psi_{\text{enc}}(T_{k,j})$ is obtained by encoding $T_{k,j}$ with the same pre-trained encoder. The explicit instruction guides the reasoning branch, while the dense vector conditions the statistical fitting branch, ensuring both processes operate in a shared semantic space.

\subsection{Policy-Guided Dual-Process Simulation}
\label{sec:dual_process}

We deploy two parallel branches, both anchored by the shared policy representations: a \textit{semantic reasoning} branch that leverages explicit decision logic via an LLM, and a \textit{statistical fitting} branch that captures implicit regularities via a supervised model.

\subsubsection*{Semantic Reasoning}
Given a scene $S$ and a sampled policy instruction $T_{k,j}$, an LLM-based agent predicts the binary decision:
\begin{equation}
    \hat{y}_{\text{reason}} = f_{\text{LLM}}(S \mid \text{SystemPrompt} = T_{k,j}).
\end{equation}
The policy text grounds LLM inference in domain-specific trade-off logic, preventing over-rationalization.

\subsubsection*{Statistical Fitting}
A supervised ML model (i.e., Gradient Boosting) captures implicit environmental regularities such as geographic convenience and offline brand perception. Unlike standard models, ours is policy-conditioned. We concatenate scene features $\phi(\tilde{S})$, user profile embedding $\mathbf{e}_{\text{prof}}$, and the policy vector $v_{k,j}$:
\begin{equation}
    \hat{y}_{\text{fit}} = f_{\text{ML}}\big(\text{Concat}( \phi(\tilde{S}), \, \mathbf{e}_{\text{prof}}, \, v_{k,j} )\big).
\end{equation}
Injecting the policy vector enables the model to learn $P(Y \mid S, \text{Policy})$, capturing interactions between decision mechanisms and scene features that the text-based reasoning branch cannot observe.

\subsection{Group-Level Aggregation}
\label{sec:fusion}

We estimate the policy distribution $\hat{P}_U(u \mid S)$ for each scene using its historical visitor log. Each visitor is assigned to the most similar policy via cosine similarity between their embedding and the policy vectors $\{v_{k,j}\}$.

We then conduct Monte Carlo simulation with $N$ samples. For each step $i$, we draw a policy $u^{(i)} \sim \hat{P}_U(u \mid S)$ and obtain predictions from both branches. The final group-level estimate fuses the two:
\begin{equation}
    \hat{Y}_{\text{hybrid}}(S, a) = \lambda \cdot \frac{1}{N} \sum_{i=1}^N \hat{y}_{\text{reason}}^{(i)} + (1-\lambda) \cdot \frac{1}{N} \sum_{i=1}^N \hat{y}_{\text{fit}}^{(i)},
\end{equation}
where $\lambda \in [0, 1]$ balances the two branches. This fusion exploits complementary error structures. The reasoning branch generalizes better under data sparsity while the fitting branch absorbs implicit regularities. Empirically, balanced fusion yields the lowest error.

\section{Experiments}
\label{sec:experiments}

\begin{table*}[t]
    \centering
    \tiny
    \caption{Baseline Comparison: Group Simulation Error (GSE\%) across merchant tiers and categories. ML models achieve lower variance but struggle with semantic generalization; LLMs exhibit higher variance due to over-rationality. }
    \label{tab:baseline_results}
    
    \definecolor{tablerowalt}{gray}{0.93}
    \definecolor{green1}{RGB}{230,245,230}
    \definecolor{green2}{RGB}{204,235,204}
    \definecolor{green3}{RGB}{179,226,179}
    \definecolor{green4}{RGB}{153,216,153}
    \definecolor{green5}{RGB}{128,207,128}
    \definecolor{green6}{RGB}{102,197,102}
    \definecolor{green7}{RGB}{77,187,77}
    
    \resizebox{0.95\linewidth}{!}{
        \begin{tabular}{l l|ccc|ccccc|cc}
            \toprule
            & & \multicolumn{3}{c|}{\textbf{Merchant Tiers (GSE \%) $\downarrow$ }} & \multicolumn{5}{c|}{\textbf{Merchant Categories (GSE \%) $\downarrow$ }} & \multicolumn{2}{c}{\textbf{Overall $\downarrow$ }} \\
            \cmidrule(lr){3-5} \cmidrule(lr){6-10} \cmidrule(lr){11-12}
            \textbf{Category} & \textbf{Model} & \textbf{Head} & \textbf{Mid} & \textbf{Tail} & \textbf{Chinese} & \textbf{Snack} & \textbf{Exotic} & \textbf{Hotpot} & \textbf{BBQ} & \textbf{GSE} (\%) & \textbf{Std} \\
            \midrule
            
            \multirow{4}{*}{\rotatebox[origin=c]{90}{\makecell{Machine\\Learning}}}
            & Logistic Regression & \cellcolor{tablerowalt}16.20 & \cellcolor{tablerowalt}17.80 & \cellcolor{tablerowalt}19.50 & \cellcolor{tablerowalt}18.20 & \cellcolor{tablerowalt}17.50 & \cellcolor{tablerowalt}15.80 & \cellcolor{tablerowalt}17.60 & \cellcolor{tablerowalt}16.90 & \cellcolor{green1}17.80 & 11.20 \\
            & XGBoost & 14.80 & 14.97 & 15.86 & 16.95 & 15.06 & 11.10 & 16.27 & 15.11 & \cellcolor{green2}15.21 & 9.50 \\
            & DNN (Embedding) & \cellcolor{tablerowalt}14.20 & \cellcolor{tablerowalt}15.10 & \cellcolor{tablerowalt}16.80 & \cellcolor{tablerowalt}15.80 & \cellcolor{tablerowalt}14.20 & \cellcolor{tablerowalt}12.50 & \cellcolor{tablerowalt}15.30 & \cellcolor{tablerowalt}13.80 & \cellcolor{green2}14.90 & 9.80 \\
            & Gradient Boosting & 14.50 & 14.80 & 15.60 & 16.70 & 14.80 & 10.80 & 16.00 & 14.90 & \cellcolor{green2}14.98 & 9.20 \\
            \midrule
            
            \multirow{4}{*}{\rotatebox[origin=c]{90}{\makecell{Zero-shot\\LLM}}}
            & Qwen3-32B & \cellcolor{tablerowalt}15.80 & \cellcolor{tablerowalt}21.10 & \cellcolor{tablerowalt}20.15 & \cellcolor{tablerowalt}21.45 & \cellcolor{tablerowalt}21.30 & \cellcolor{tablerowalt}17.55 & \cellcolor{tablerowalt}18.20 & \cellcolor{tablerowalt}14.65 & \cellcolor{green1}18.95 & 13.20 \\
            & DeepSeek-V3 & 15.20 & 20.45 & 19.65 & 20.90 & 20.75 & 16.90 & 17.55 & 14.05 & \cellcolor{green1}18.42 & 12.85 \\
            & GPT-4o & \cellcolor{tablerowalt}14.75 & \cellcolor{tablerowalt}20.67 & \cellcolor{tablerowalt}19.41 & \cellcolor{tablerowalt}20.79 & \cellcolor{tablerowalt}20.99 & \cellcolor{tablerowalt}16.61 & \cellcolor{tablerowalt}17.28 & \cellcolor{tablerowalt}13.76 & \cellcolor{green1}18.27 & 13.03 \\
            & GPT-4.1 & 14.20 & 19.80 & 18.90 & 20.10 & 19.50 & 16.20 & 16.80 & 13.20 & \cellcolor{green2}17.65 & 12.50 \\
            \midrule
            
            \multirow{4}{*}{\rotatebox[origin=c]{90}{\makecell{Few-shot\\LLM}}}
            & Qwen3-32B & \cellcolor{tablerowalt}14.65 & \cellcolor{tablerowalt}19.20 & \cellcolor{tablerowalt}18.85 & \cellcolor{tablerowalt}19.80 & \cellcolor{tablerowalt}18.65 & \cellcolor{tablerowalt}16.45 & \cellcolor{tablerowalt}17.10 & \cellcolor{tablerowalt}13.95 & \cellcolor{green2}17.55 & 12.35 \\
            & DeepSeek-V3 & 13.95 & 18.55 & 18.15 & 19.20 & 17.95 & 15.80 & 16.50 & 13.55 & \cellcolor{green2}16.98 & 11.95 \\
            & GPT-4o & \cellcolor{tablerowalt}13.80 & \cellcolor{tablerowalt}18.20 & \cellcolor{tablerowalt}17.80 & \cellcolor{tablerowalt}18.90 & \cellcolor{tablerowalt}17.80 & \cellcolor{tablerowalt}15.50 & \cellcolor{tablerowalt}16.20 & \cellcolor{tablerowalt}13.50 & \cellcolor{green2}16.80 & 11.80 \\
            & GPT-4.1 & 13.35 & 17.65 & 17.20 & 18.35 & 17.25 & 15.05 & 15.75 & 12.95 & \cellcolor{green3}16.25 & 11.45 \\
            \midrule
            
            \multirow{2}{*}{\rotatebox[origin=c]{0}{\textbf{Ours}}}
            & \textbf{PGHS (w/o Policy)} & \cellcolor{tablerowalt}8.32 & \cellcolor{tablerowalt}10.43 & \cellcolor{tablerowalt}11.83 & \cellcolor{tablerowalt}10.13 & \cellcolor{tablerowalt}10.16 & \cellcolor{tablerowalt}10.44 & \cellcolor{tablerowalt}9.68 & \cellcolor{tablerowalt}9.16 & \cellcolor{green3}9.91 & 7.37 \\
            & \textbf{PGHS (Full)} & \textbf{7.57} & \textbf{7.99} & 12.26 & \textbf{8.87} & 10.29 & \textbf{8.22} & \textbf{8.29} & \textbf{8.24} & \cellcolor{green4}\textbf{8.80} & \textbf{6.62} \\
            \bottomrule
        \end{tabular}
    }
\end{table*}

\subsection{Experimental Setup}
\label{sec:setup}
\subsubsection*{Datasets}
\label{sec:datasets}
\begin{figure}[t]
    \centering
    \includegraphics[width=\linewidth]{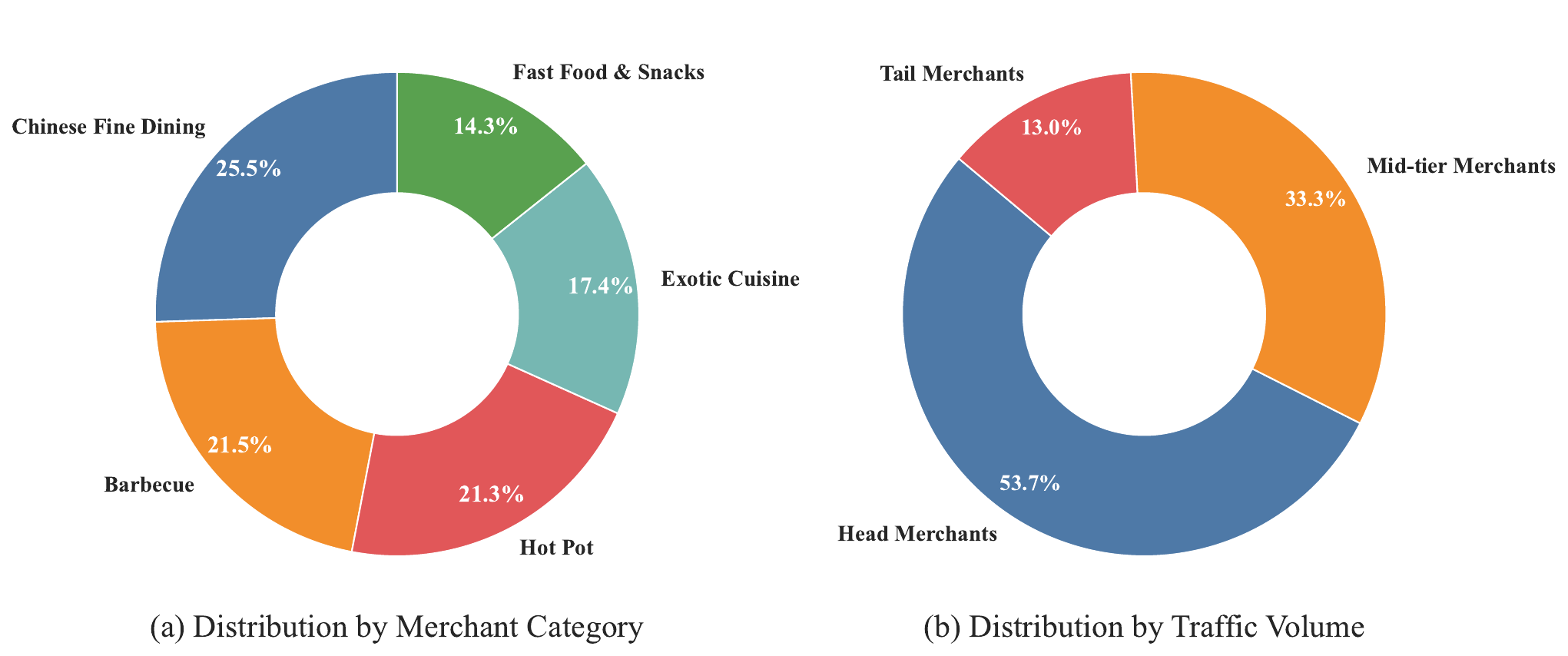}
    \caption{Statistical distribution of the dataset by merchant category (a) and traffic volume tier (b).}
    \label{fig:data_distribution}
\end{figure}

We evaluate PGHS on real-world production data from the Meituan platform, sampled from online user interaction logs during October to November 2025. The evaluation set covers 26,461 user decision trajectories across 101 merchants, spanning five culinary categories and three traffic tiers. 
Each trajectory records a complete search-to-purchase sequence with explicit comparison actions. User profiles include demographics, consumption habits, and taste preferences. Merchant profiles include category, price tier, rating, and promotional status. All profiles are encoded into dense embeddings via a pre-trained encoder. 
We adopt a chronological split where the first 20\% of trajectories are used for policy mining, and the remaining 80\% serve as the test set. Figure~\ref{fig:data_distribution} shows the distribution of these trajectories across different merchant categories and traffic volume tiers.

\subsubsection*{Baselines}
We compare PGHS against two categories of baselines. 
\textit{Fitting-based (ML)} baselines predict binary purchase decisions from concatenated user and merchant embeddings. We evaluate \texttt{Logistic Regression}, \texttt{XGBoost}, \texttt{Gradient Boosting}, and a 3-layer \texttt{DNN}. 
\textit{Reasoning-based (LLM)} baselines predict purchase decisions from textual descriptions. We evaluate \texttt{GPT-4o}, \texttt{GPT-4.1}, \texttt{DeepSeek-V3}, and \texttt{Qwen3-32B} in both Zero-shot and Few-shot settings. 

\subsubsection*{Metrics}
We evaluate simulation fidelity using Group Simulation Error (GSE), defined as the mean absolute error between predicted and ground-truth per-merchant choice rates. We also report GSE Standard Deviation (GSE-SD) to measure prediction stability across heterogeneous merchant types. Lower values indicate better accuracy and robustness.

\subsection{Overall Simulation Performance}
\label{sec:overall}

Table~\ref{tab:baseline_results} reports GSE across merchant tiers and categories. Among ML baselines, DNN (Embedding) achieves the lowest error (14.90\%) while Gradient Boosting exhibits the most stable cross-tier performance, though all ML models degrade on Tail merchants due to data sparsity. Among LLM baselines, GPT-4.1 Few-shot achieves the best error (16.25\%), but all LLMs exhibit higher variance than ML models due to over-rationalization on implicit-dominant categories such as Snack. PGHS achieves a GSE of 8.80\% with a standard deviation of 6.62\%, representing relative improvements of 45.8\% over the best LLM baseline and 40.9\% over the best ML baseline.

\subsection{Ablation Study}
\label{sec:ablation}

\begin{figure}[t]
    \centering
    \includegraphics[width=\linewidth]{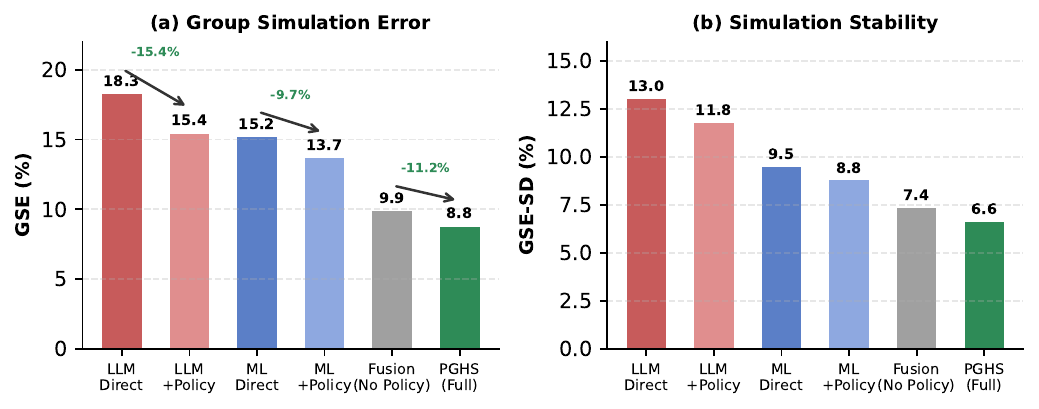}
    \caption{Ablation study on policy guidance and dual-process fusion.}
    \label{fig:ablation}
\end{figure}

Figure~\ref{fig:ablation} compares GSE and GSE-SD across six model variants to isolate the contributions of policy guidance and dual-process fusion. 
Policy guidance reduces GSE by 15.4\% for the LLM branch and 9.7\% for the ML branch. Dual-process fusion further amplifies the gains. Even without policy guidance, fusion alone already outperforms the best single-branch model. With policy guidance, PGHS (Full) achieves 8.80\% by exploiting complementary error structures.

We also compare three variants of the policy discovery pipeline. Without clustering, each user is treated as a unique policy, leading to overfitting and a 23\% GSE increase. Using K-Means alone captures coarse persona groups but fails to isolate distinct strategies within each group, resulting in a 12\% GSE increase. The full hierarchical pipeline (K-Means + HDBSCAN) yields the best results.

\subsection{Expert Evaluation for Business Diagnosis}
\label{sec:expert_eval}

PGHS has been developed and piloted within Meituan as a merchant diagnostic tool. To validate its practical effectiveness, we conduct an expert evaluation. We select 10 merchants spanning diverse categories and tiers, and define 8 candidate operational strategies (e.g., discount promotions, combo meal additions, delivery fee adjustments, storefront redesign). Three senior operations experts independently rank the strategies per merchant by expected effectiveness, and the consensus ranking serves as gold standard. Each model simulates the 8 strategies per merchant and ranks them by predicted conversion rate. We measure ranking quality using Kendall's $\tau$, which quantifies ordinal agreement with expert consensus ($\tau \in [-1, 1]$, higher is better).

\begin{table}[t]
    \centering
    \small
    \caption{Kendall's $\tau$ between model-predicted and expert strategy rankings, averaged over 10 merchants.}
    \label{tab:expert_eval}
    \begin{tabular}{lc}
        \toprule
        \textbf{Model} & \textbf{Kendall's $\tau$ $\uparrow$} \\
        \midrule
        XGBoost & 0.45 \\
        Gradient Boosting & 0.50 \\
        GPT-4.1 (Few-shot) & 0.52 \\
        DeepSeek-V3 (Few-shot) & 0.50 \\
        PGHS (w/o Policy) & 0.62 \\
        \textbf{PGHS (Full)} & \textbf{0.69} \\
        \bottomrule
    \end{tabular}
\end{table}

As shown in Table~\ref{tab:expert_eval}, PGHS (Full) achieves a Kendall's $\tau$ of 0.69, outperforming the best ML baseline by 38.0\% and the best LLM baseline by 32.7\%. This confirms that PGHS produces strategy rankings well-aligned with expert judgment, validating its utility for real-world merchant diagnosis.

\section{Conclusion}
We presented PGHS, a policy-guided dual-process framework for group-level user simulation. Deployed on Meituan with 101 merchants and 26k+ trajectories, PGHS achieves 8.80\% group simulation error, improving over the best reasoning-based and fitting-based baselines by 45.8\% and 40.9\% respectively. While our evaluation focuses on merchant-level diagnosis, the dual-process architecture naturally extends to item-level recommendation scenarios where group behavior prediction is required. Future work includes adaptive fusion mechanisms, temporal policy dynamics, and validation on public e-commerce benchmarks.

\begin{acks}
This work was partially supported by 
National Natural Science Foundation of China (Nos. U25B2047, 62272469, 72301284), and
The Science and Technology Innovation Program of Hunan Province (No. 2023RC1007).
\end{acks}

\section*{Presenter and Company Bio}
Ziyang Chen is a research intern in the LongCat Interaction Team at Meituan Inc., focusing on social simulation and LLM-based decision-making. He is currently pursuing his PhD at the National University of Defense Technology after receiving his bachelor's degree in Information and Computing Science.

Meituan is China's leading local services platform, connecting hundreds of millions of consumers with merchants across food delivery, hotel and travel booking, in-store dining, and everyday lifestyle services.

\bibliographystyle{ACM-Reference-Format}
\balance
\bibliography{sample-base}

\end{document}